\pdfoutput=1

\documentclass[11pt]{article}
\usepackage{amssymb}

\usepackage[preprint]{acl}
\usepackage[many]{tcolorbox}
\usepackage{times}
\usepackage{array}
\usepackage{pifont}
\usepackage{amssymb}
\usepackage{pifont}
\newcommand{\cmark}{\ding{51}}  
\usepackage{latexsym}
\usepackage{xcolor}
\usepackage{tcolorbox}
\usepackage{listings}
\usepackage{tabularx}
\usepackage{booktabs}
\usepackage{enumitem}
\usepackage[T1]{fontenc}

\usepackage[utf8]{inputenc}

\usepackage{microtype}
\usepackage{amsmath}
\usepackage{amssymb}
\usepackage{colortbl}
\usepackage{booktabs}

\usepackage{xcolor}
\usepackage{multirow}
\usepackage{inconsolata}
\usepackage{graphicx}
\hypersetup{
    colorlinks=true,
    linkcolor=red,
    citecolor=cyan,
    filecolor=magenta,      
    urlcolor=cyan,
    }
\title{
TAGS: A Test-Time Generalist–Specialist Framework with Retrieval-Augmented Reasoning and Verification
}

\author{
Jianghao Wu$^{1, 2}$, 
Feilong Tang$^{1, 2}$, 
Yulong Li$^{1, 3}$,
Ming Hu$^{1,2}$,
Haochen Xue$^{1, 3}$,\\
\textbf{Shoaib Jameel$^{4}$, }
\textbf{Yutong Xie$^{1}$, }
\textbf{Imran Razzak}$^{1\dag}$ \\
$^1$Mohamed bin Zayed University of Artificial Intelligence \quad
$^2$Monash University \quad  \\
$^3$Xi’an Jiaotong-Liverpool University \quad 
$^4$University of Southampton \quad \\
\texttt{Imran.razzak@mbzuai.ac.ae}
}

\begin{document}

\maketitle
\renewcommand{\thefootnote}{\dag}
\footnotetext{Corresponding author.}

\begin{abstract}
Recent advances such as Chain-of-Thought prompting have significantly improved large language models (LLMs) in zero-shot medical reasoning. However, prompting-based methods often remain shallow and unstable, while fine-tuned medical LLMs suffer from poor generalization under distribution shifts and limited adaptability to unseen clinical scenarios.
To address these limitations, we present \textbf{TAGS}, a test-time framework that combines a broadly capable generalist with a domain-specific specialist to offer complementary perspectives without any model fine-tuning or parameter updates.
To support this generalist–specialist reasoning process, we introduce two auxiliary modules: a hierarchical retrieval mechanism that provides multi-scale exemplars by selecting examples based on both semantic and rationale-level similarity, and a reliability scorer that evaluates reasoning consistency to guide final answer aggregation.
TAGS achieves strong performance across nine MedQA benchmarks, boosting GPT-4o accuracy by 13.8\%, DeepSeek-R1 by 16.8\%, and improving a vanilla 7B model from 14.1\% to 23.9\%. These results surpass several fine-tuned medical LLMs, without any parameter updates. The code will be available at \url{https://github.com/JianghaoWu/TAGS}.
\end{abstract}

\section{Introduction}
Large Language Models (LLMs) have recently demonstrated promising capabilities in medical question answering (MedQA), achieving strong performance on a range of benchmark datasets~\cite{singhal2025toward, jin2022biomedical, chen2023large, chen2023systematic, zhou2023survey, gao2024empowering}. However, despite these advances, recent studies have shown that even state-of-the-art models frequently fail on complex cases requiring deep domain knowledge, multi-step reasoning, and generalization to out-of-distribution clinical scenarios~\cite{xu2024towards, fan2025ai, shi2024medadapter}. 

To improve LLM reasoning in MedQA, two major research directions have emerged. The first involves prompting-based strategies, such as Chain-of-Thought (CoT)~\cite{wei2022chain} and Multi-Agent Systems (MAS)~\cite{chen2025enhancing, chen2025mdteamgpt}, which guide LLMs through structured multi-step reasoning or simulate expert collaboration. 
However, empirical studies~\cite{tang2025medagentsbench} have revealed that interactive multi-agent reasoning is frequently brittle, with redundant outputs, unstable communication, and poor coordination undermining reliability. Recent benchmarks report high failure rates for such systems, with multi-agent discussions yielding limited improvements over single-agent baselines~\cite{cemri2025multi}.
The second direction focuses on fine-tuning LLMs on domain-specific medical data, yielding specialist models like \textsc{HuatuoGPT}~\cite{chen2024huatuogpt} and \textsc{MedLLaMA}~\cite{qiu2024towards}. Although these models perform well on in-distribution tasks, they tend to overfit their training domains and exhibit limited adaptability to emerging knowledge or unseen questions~\cite{yang-etal-2024-self, ye2024gmai}.

While prior work has made progress on reasoning, retrieval, and domain adaptation, these components are typically developed in isolation and lack integration into a unified, inference-time framework. 
As a result, existing approaches fall short in three critical aspects for robust MedQA: acquiring up-to-date medical knowledge, supporting diverse and complementary reasoning strategies, and ensuring answer reliability under distribution shift, all without relying on any parameter updates.
This motivates our work: designing a structured, test-time reasoning framework that unifies generalist and specialist perspectives with retrieval-augmented prompting and verification.

Our motivation stems from the observation that single-agent prompting often lacks the depth and diversity required for complex medical reasoning, while multi-agent systems tend to produce redundant or inconsistent outputs due to unclear role assignments and weak complementarity. To address these limitations, we propose \textbf{TAGS} (Test-time Generalist–Specialist framework with retrieval-augmented reasoning and verification), a modular, inference-only framework that enhances medical question answering through structured reasoning collaboration. At its core, TAGS relies on a Generalist–Specialist Reasoning Collaboration (\textbf{GSRC}) module, which facilitates complementary reasoning between a generalist agent and a specialist agent. Each agent brings a different perspective to the problem: the generalist offers broad clinical insights, while the specialist contributes precise, domain-specific reasoning. This collaboration is designed to produce diverse and accurate reasoning paths for challenging medical questions.

To support GSRC, we introduce two auxiliary modules. First, to promote diverse and contextually grounded reasoning, we propose a Hierarchical Retrieval Augmentation (\textbf{HRA}) mechanism that leverages external, high-quality medical Chain-of-Thought (CoT) exemplars. HRA operates in two stages: it first retrieves semantically relevant CoT exemplars based on the input question–answer (QA) pair to serve as initial references for both agents. Then, after the agents generate their initial reasoning steps, HRA performs a second round of retrieval based on the generated rationales to obtain reasoning-level exemplars that further guide and enrich subsequent inference.
Second, to ensure the reliability of the generated answers along with their reasoning chains, we introduce an Uncertainty-Aware Answer Aggregation (\textbf{UAAA}) module. UAAA evaluates the consistency between the reasoning path and the final answer produced by each agent, and aggregates the outputs based on a confidence score and inter-agent agreement, yielding a robust and interpretable final prediction.
This design effectively mitigates issues common in previous methods, such as insufficient reasoning depth and diversity in single-agent prompting and redundant or conflicting outputs arising from ambiguous roles in multi-agent systems, ultimately producing more accurate and robust medical reasoning outcomes.

Extensive experiments across nine MedQA benchmarks validate the robustness and adaptability of \textsc{TAGS} under both non-fine-tuned and fine-tuned settings. Our method consistently outperforms strong prompting and agent-based baselines across multiple foundation models, including GPT-4o (18.0\% to 31.8\%), DeepSeek-R1 (27.2\% to 44.0\%), and Qwen-2.5-7B (14.1\% to 23.9\%). Notably, we show that even without retrieving semantically similar exemplars, \textsc{TAGS} maintains strong performance by exposing models to diverse reasoning patterns rather than specific answers.

\begin{figure*}[t]
\centering
\includegraphics[width=\textwidth]{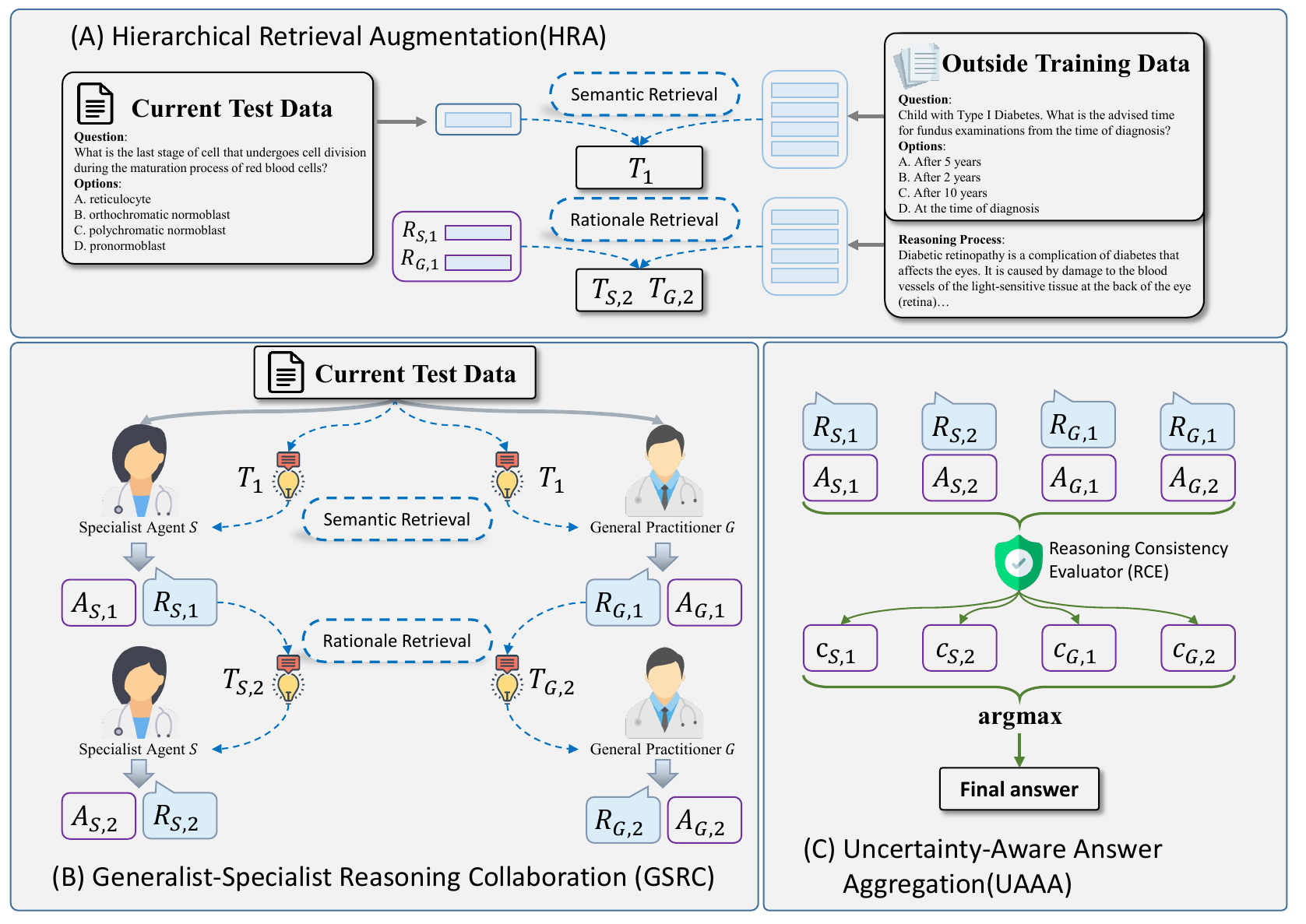}
\caption{
    Overview of the proposed \textsc{TAGS} framework. The architecture consists of three modules: 
    (A) \textbf{HRA (Hierarchical Retrieval Augmentation)}, a two-stage retrieval process that selects semantically relevant exemplars (\(\mathcal{T}_1\)) and refines them based on rationale alignment (\(\mathcal{T}_{G,2}\), \(\mathcal{T}_{S,2}\)). 
    (B) \textbf{GSRC (Generalist-Specialist Reasoning Collaboration)} employs dual-agent reasoning across two rounds, generating four candidate (Rationale, Answer) pairs. 
    (C) \textbf{UAAA (Uncertainty-Aware Answer Aggregation)} assesses rationale consistency using the RCE and aggregates reliability scores (\(c\)) to determine the final answer.
}
\label{fig_pipeline}
\end{figure*}

\section{Related Work}
\vspace{-0.5em}

\subsection{Medical Question Answering}
Medical question answering (MedQA) aims to predict accurate answers to domain-specific clinical or biomedical questions, often posed in a multiple-choice format. Existing benchmarks span a variety of formats and reasoning challenges, including clinical exam-style questions, evidence-based inference, and multi-subject distractor-rich scenarios~\cite{pal2022medmcqa}.
Recent approaches leverage large language models (LLMs) or chain-of-thought prompting to enhance reasoning~\cite{singhal2025toward}. MedCoT~\cite{liu2024medcot} explicitly integrates multi-step rationale generation with hierarchical expert feedback. Concurrently, biomedical LLMs such as MedLLaMA~\footnote{\url{https://huggingface.co/johnsnowlabs/}}, HuatuoGPT~\cite{chen2024huatuogpt}, and OpenBioLLM~\cite{pal2024openbiollms} have achieved strong zero-shot or few-shot performance on MedQA benchmarks. However, these models typically rely on direct answer generation and lack explicit mechanisms for multi-agent reasoning or consistency verification.
In contrast, our method introduces a retrieval-augmented multi-agent framework that performs staged reasoning and employs a dedicated verifier to assess the reliability of generated answers, promoting robustness and interpretability.

\subsection{Retrieval-Augmented Reasoning}

Retrieval-augmented reasoning enhances prediction quality, factual consistency, and interpretability by incorporating external knowledge into the reasoning process. Early works such as RAG~\cite{lewis2020retrieval} retrieve text passages to guide open-domain generation, while later methods extend retrieval to more structured forms, such as few-shot demonstrations~\cite{izacard2023atlas} or intermediate reasoning paths~\cite{shi2023replug}.
In the context of chain-of-thought (CoT) prompting, retrieval has been explored to select relevant questions, rationales, or multi-step reasoning exemplars that better align with the target task~\cite{xu2022prompting, he2025inference}. Despite these advances, most existing frameworks rely primarily on question-level similarity and often neglect deeper alignment at the reasoning level.
Our framework addresses this limitation by employing a hierarchical retrieval strategy: first retrieving question-option exemplars, then refining based on CoT similarity. This enables alignment in both problem context and reasoning structure, thereby improving downstream multi-agent reasoning.

\subsection{Multi-Agent Systems for Reasoning}
Multi-agent systems (MAS) have emerged as a promising approach to enhance the robustness, diversity, and reliability of reasoning in complex tasks, including medical question answering. By orchestrating multiple reasoning paths or personas, MAS frameworks aim to mitigate biases and capture complementary perspectives, which are particularly critical in high-stakes medical decision-making.
Recent studies have explored various MAS paradigms for medical reasoning. MedAgents~\cite{tang2023medagents} proposes a collaborative multi-agent framework where multiple agents independently generate answers and a majority voting scheme determines the final prediction. MDAgents~\cite{kim2024mdagents} further enhances this idea by introducing dynamic collaboration and adaptive feedback mechanisms among agents during the reasoning process. MedPrompt~\cite{chen2024medprompt} adopts a multi-round prompting strategy combined with ensemble voting to improve medical QA performance. Additionally, frameworks like Multi-Persona~\cite{wang2023multipersona} and Self-Refine~\cite{madaan2024selfrefine} leverage self-collaboration and iterative self-feedback to strengthen individual agent reasoning capabilities.
While multi-agent collaboration has demonstrated effectiveness in improving answer quality, it also introduces notable challenges. As highlighted in recent evaluations~\cite{tang2023medagents}, excessive agent interactions may lead to reasoning conflicts, unstable decision paths, and increased inference costs. Recent studies~\cite{cemri2025multi} further reveal that over-complex MAS architectures often suffer from systemic failures, such as miscommunication, vague role specification, and weak verification. Moreover, most existing MAS frameworks lack explicit mechanisms to assess the internal consistency between generated reasoning and final answers, which can limit reliability in clinical contexts.
To address these limitations, we propose a lightweight General-Specialist Reasoning Collaboration (GSRC) strategy that pairs a generalist and a specialist agent in a complementary manner, promoting stable and robust medical reasoning with minimal inter-agent conflicts.

\section{Methodology}
\label{sec:method}
We propose \textsc{TAGS} (Test-time Generalist–Specialist Reasoning with Retrieval-Augmentation and Uncertainty-Aware Verification), a parameter-efficient framework for medical question answering that operates entirely during inference. At its core is the Generalist–Specialist Reasoning Collaboration (GSRC), a dual-agent design that promotes reasoning diversity and domain alignment without requiring any parameter updates. To support GSRC, we introduce two auxiliary modules: Hierarchical Retrieval Augmentation (HRA), which supplies diverse and rationale-aligned exemplars, and Uncertainty-Aware Answer Aggregation (UAAA), which selects the final answer by evaluating the consistency of each reasoning path. As shown in Figure~\ref{fig_pipeline}, these components form an integrated pipeline that enables robust, zero-shot clinical QA without model fine-tuning.

\subsection{Hierarchical Retrieval Augmentation}
\label{subsec:hra}
Hierarchical Retrieval Augmentation (HRA) grounds reasoning in up-to-date evidence while injecting diverse paths for chain-of-thought (CoT) generation through a two-stage retrieval scheme. We retrieve from a frozen medical-QA corpus $\mathcal{D}$ whose entries are $d_i=(Q_i,O_i,A_i,R_i)$, where $R_i$ denotes the CoT rationale. We use a frozen text encoder $\mathcal{E}(\cdot)$ based on M3-Embedding~\cite{chen2024bge}, with 1024-dimensional output.

\textbf{Stage 1: Initial semantic retrieval.}  
We begin by embedding the query using a frozen encoder. Let $\mathbf{z} = \mathcal{E}(Q \oplus O)$, where the question $Q$ and its options $O$ are concatenated in standard order (A, B, C, D). Cosine similarity is computed against all candidate embeddings $\mathcal{E}(Q_i \oplus O_i)$ in the corpus. The top-$K$ retrieved examples form:
\begin{equation}
\label{eq:t1}
\mathcal{T}_1 = \operatorname*{Top\text{-}K} \bigl\{ d_i \in \mathcal{D} : \mathrm{sim}(\mathbf{z}, \mathcal{E}(Q_i \oplus O_i)) \bigr\}.
\end{equation}

\textbf{Stage 2: Rationale-guided retrieval.}  
After Round-1 reasoning yields preliminary rationales $R_{G,1}$ and $R_{S,1}$, we retrieve exemplars whose stored rationales best match these CoTs:
\begin{equation}
\label{eq:tg2}
\mathcal{T}_{G,2} = \operatorname*{Top\text{-}K} \bigl\{ d_i \in \mathcal{D} : \mathrm{sim}(\mathbf{r}_G, \mathcal{E}(R_i)) \bigr\},
\end{equation}
\begin{equation}
\label{eq:ts2}
\mathcal{T}_{S,2} = \operatorname*{Top\text{-}K} \bigl\{ d_i \in \mathcal{D} : \mathrm{sim}(\mathbf{r}_S, \mathcal{E}(R_i)) \bigr\}.
\end{equation}
By aligning on reasoning paths rather than surface form, Stage 2 injects complementary evidence beyond surface similarity, reducing the limitations of purely semantic matching.

\subsection{Generalist–Specialist Collaboration}
\label{subsec:gsrc}
Given the retrieved exemplar sets from HRA, Generalist–Specialist Reasoning Collaboration (GSRC) performs dual-agent inference in two rounds by coupling broad medical knowledge with focused domain expertise. The system consists of a generalist agent $\mathcal{G}$ and a specialist agent $\mathcal{S}$, both instantiated as prompted roles of the same frozen LLM without parameter updates.

An auxiliary LLM role first infers the medical specialty most relevant to the query $(Q,O)$, yielding a label $s$ (e.g., \texttt{cardiology}). This label is then injected into the prompt for $\mathcal{S}$ as “You are a medical specialist in the field of [s]”, guiding its reasoning toward domain-specific knowledge while preserving the core semantics of $(Q,O)$. Further details are provided in Appendix~\ref{Generalist–Specialist System Roles and Prompt}.
The collaboration unfolds in two rounds that iteratively refine rationales and answers.

\textbf{Round 1: Initial hypothesis generation.}  
Both agents receive the query $(Q,O)$ together with the semantically retrieved set
$\mathcal{T}_{1}$ (§\ref{subsec:hra}); the specialist additionally sees the inferred
specialty $s$.  Each agent produces an initial CoT and answer:
\begin{equation}
\label{eq:gsrc_r1}
\begin{aligned}
\bigl(R_{G,1},A_{G,1}\bigr) &= \mathcal{G}\!\bigl(Q,O,\mathcal{T}_{1}\bigr),\\
\bigl(R_{S,1},A_{S,1}\bigr) &= \mathcal{S}\!\bigl(Q,O,\mathcal{T}_{1},s\bigr).
\end{aligned}
\end{equation}
These preliminary CoTs trigger Stage 2 of HRA, which returns the rationale-aligned
exemplar sets $\mathcal{T}_{G,2}$ and $\mathcal{T}_{S,2}$.

\textbf{Round 2: Refined reasoning with aligned exemplars.}  
Using the tailored sets, the agents generate updated rationales and answers:
\begin{equation}
\label{eq:gsrc_r2}
\begin{aligned}
\bigl(R_{G,2},A_{G,2}\bigr) &= \mathcal{G}\!\bigl(Q,O,\mathcal{T}_{G,2}\bigr),\\
\bigl(R_{S,2},A_{S,2}\bigr) &= \mathcal{S}\!\bigl(Q,O,\mathcal{T}_{S,2},s\bigr).
\end{aligned}
\end{equation}

\noindent
Finally, the four \emph{(rationale, answer)} pairs are gathered into a candidate set
{\small
\begin{equation}
\label{eq:candidates}
\mathcal{C}
  = \bigl\{\, (R_{k,r},A_{k,r}) \mid k\!\in\!\{G,S\},\; r\!\in\!\{1,2\} \bigr\},
\end{equation}}
which is then forwarded to the Uncertainty-Aware Answer Aggregation module
(§\ref{subsec:uaaa}) for scoring and final selection.

\subsection{Uncertainty-Aware Answer Aggregation}
\label{subsec:uaaa}

Uncertainty-Aware Answer Aggregation (UAAA) takes as input the candidate set $\mathcal{C}$ generated by GSRC (§\ref{subsec:gsrc}) and selects a single high-confidence answer through consistency-based scoring. To accomplish this, we define a \emph{Reasoning Consistency Evaluator} (RCE), implemented as a separate zero-shot role of the same frozen LLM.

Given a candidate pair $(R_k, A_k)$, the RCE assesses how well the rationale supports the answer in the context of the original query $(Q, O)$, and assigns an integer score $c_k \in [0, 5]$, where higher values indicate stronger logical and clinical coherence. The scoring rubric is detailed in Appendix~\ref{A_Reasoning Consistency Evaluator Rubri}.
The final answer is selected as:
\begin{equation}
\label{eq:uaaa_select}
A_{\text{final}} = A_{k^\star},
\quad
k^\star = \underset{k \in \mathcal{C}}{\arg\max}\; c_k.
\end{equation}

In the case of ties, preference is resolved deterministically in the following order: specialist round 2, generalist round 2, specialist round 1, and generalist round 1. By explicitly verifying the internal consistency of each reasoning path, UAAA mitigates hallucination propagation and stabilizes final predictions, all without any parameter updates.

\begin{table*}[htbp]
  \centering

  \resizebox{\textwidth}{!}{
\begin{tabular}{l|c|c|c|c|c|c|c|c|c|c}
\toprule
Method & MedQA & PubMedQA & MedMCQA & MedBullets & MMLU & MMLU-Pro & MedExQA & MedXpert-R & MedXpert-U & Average \\
\midrule
GPT-4o & \cellcolor[rgb]{0.951,0.981,0.942}32.0 & \cellcolor[rgb]{0.951,0.981,0.942}9.0 & \cellcolor[rgb]{0.951,0.981,0.942}25.0 & \cellcolor[rgb]{0.951,0.981,0.942}19.1 & \cellcolor[rgb]{0.951,0.981,0.942}24.7 & \cellcolor[rgb]{0.951,0.981,0.942}21.0 & \cellcolor[rgb]{0.754,0.897,0.753}18.0 & \cellcolor[rgb]{0.951,0.981,0.942}7.0 & \cellcolor[rgb]{0.951,0.981,0.942}6.0 & \cellcolor[rgb]{0.951,0.981,0.942}18.0 \\
+ few-shot & \cellcolor[rgb]{0.984,0.994,0.980}31.0 & \cellcolor[rgb]{0.500,0.633,0.553}16.0 & \cellcolor[rgb]{0.558,0.764,0.628}34.0 & \cellcolor[rgb]{0.984,0.994,0.980}16.9 & \cellcolor[rgb]{0.635,0.839,0.686}32.9 & \cellcolor[rgb]{0.878,0.952,0.865}27.0 & \cellcolor[rgb]{0.878,0.952,0.865}17.0 & \cellcolor[rgb]{0.878,0.952,0.865}8.0 & \cellcolor[rgb]{0.754,0.897,0.753}11.0 & \cellcolor[rgb]{0.878,0.952,0.865}21.7 \\
+ RAG & \cellcolor[rgb]{0.635,0.839,0.686}42.0 & \cellcolor[rgb]{0.754,0.897,0.753}12.0 & \cellcolor[rgb]{0.754,0.897,0.753}30.0 & \cellcolor[rgb]{0.878,0.952,0.865}22.5 & \cellcolor[rgb]{0.984,0.994,0.980}20.5 & \cellcolor[rgb]{0.635,0.839,0.686}37.0 & \cellcolor[rgb]{0.951,0.981,0.942}15.0 & \cellcolor[rgb]{0.558,0.764,0.628}19.0 & \cellcolor[rgb]{0.878,0.952,0.865}10.0 & \cellcolor[rgb]{0.754,0.897,0.753}23.1 \\
+ CoT & \cellcolor[rgb]{0.754,0.897,0.753}39.0 & \cellcolor[rgb]{0.951,0.981,0.942}10.0 & \cellcolor[rgb]{0.754,0.897,0.753}30.0 & \cellcolor[rgb]{0.635,0.839,0.686}28.1 & \cellcolor[rgb]{0.878,0.952,0.865}26.0 & \cellcolor[rgb]{0.754,0.897,0.753}35.0 & \cellcolor[rgb]{0.500,0.633,0.553}24.0 & \cellcolor[rgb]{0.754,0.897,0.753}12.0 & \cellcolor[rgb]{0.635,0.839,0.686}15.0 & \cellcolor[rgb]{0.754,0.897,0.753}24.3 \\
+ CoT-SC & \cellcolor[rgb]{0.878,0.952,0.865}37.0 & \cellcolor[rgb]{0.984,0.994,0.980}6.0 & \cellcolor[rgb]{0.500,0.633,0.553}35.0 & \cellcolor[rgb]{0.558,0.764,0.628}30.3 & \cellcolor[rgb]{0.754,0.897,0.753}30.1 & \cellcolor[rgb]{0.558,0.764,0.628}43.0 & \cellcolor[rgb]{0.558,0.764,0.628}22.0 & \cellcolor[rgb]{0.754,0.897,0.753}10.0 & \cellcolor[rgb]{0.754,0.897,0.753}14.0 & \cellcolor[rgb]{0.635,0.839,0.686}25.3 \\
+ Multi-Persona & \cellcolor[rgb]{0.558,0.764,0.628}45.0 & \cellcolor[rgb]{0.558,0.764,0.628}15.0 & \cellcolor[rgb]{0.951,0.981,0.942}25.0 & \cellcolor[rgb]{0.635,0.839,0.686}29.2 & \cellcolor[rgb]{0.558,0.764,0.628}37.0 & \cellcolor[rgb]{0.635,0.839,0.686}42.0 & \cellcolor[rgb]{0.635,0.839,0.686}21.0 & \cellcolor[rgb]{0.754,0.897,0.753}10.0 & \cellcolor[rgb]{0.635,0.839,0.686}16.0 & \cellcolor[rgb]{0.635,0.839,0.686}26.7 \\
+ Self-Refine & \cellcolor[rgb]{0.754,0.897,0.753}41.0 & \cellcolor[rgb]{0.635,0.839,0.686}13.0 & \cellcolor[rgb]{0.558,0.764,0.628}34.0 & \cellcolor[rgb]{0.635,0.839,0.686}28.1 & \cellcolor[rgb]{0.635,0.839,0.686}34.2 & \cellcolor[rgb]{0.754,0.897,0.753}34.0 & \cellcolor[rgb]{0.558,0.764,0.628}22.0 & \cellcolor[rgb]{0.635,0.839,0.686}17.0 & \cellcolor[rgb]{0.558,0.764,0.628}19.0 & \cellcolor[rgb]{0.558,0.764,0.628}26.9 \\
+ MedAgents & \cellcolor[rgb]{0.635,0.839,0.686}43.0 & \cellcolor[rgb]{0.558,0.764,0.628}15.0 & \cellcolor[rgb]{0.635,0.839,0.686}30.0 & \cellcolor[rgb]{0.754,0.897,0.753}27.0 & \cellcolor[rgb]{0.878,0.952,0.865}28.8 & \cellcolor[rgb]{0.984,0.994,0.980}8.0 & \cellcolor[rgb]{0.754,0.897,0.753}19.0 & \cellcolor[rgb]{0.984,0.994,0.980}3.0 & \cellcolor[rgb]{0.951,0.981,0.942}6.0 & \cellcolor[rgb]{0.878,0.952,0.865}20.0 \\
+ MDAgents & \cellcolor[rgb]{0.878,0.952,0.865}36.0 & \cellcolor[rgb]{0.754,0.897,0.753}11.0 & \cellcolor[rgb]{0.984,0.994,0.980}22.0 & \cellcolor[rgb]{0.878,0.952,0.865}21.3 & \cellcolor[rgb]{0.951,0.981,0.942}24.7 & \cellcolor[rgb]{0.984,0.994,0.980}8.0 & \cellcolor[rgb]{0.984,0.994,0.980}13.0 & \cellcolor[rgb]{0.951,0.981,0.942}4.0 & \cellcolor[rgb]{0.984,0.994,0.980}5.0 & \cellcolor[rgb]{0.984,0.994,0.980}16.1 \\
+ MedPrompt & \cellcolor[rgb]{0.951,0.981,0.942}34.0 & \cellcolor[rgb]{0.754,0.897,0.753}11.0 & \cellcolor[rgb]{0.878,0.952,0.865}26.0 & \cellcolor[rgb]{0.951,0.981,0.942}22.5 & \cellcolor[rgb]{0.878,0.952,0.865}26.0 & \cellcolor[rgb]{0.878,0.952,0.865}22.0 & \cellcolor[rgb]{0.878,0.952,0.865}16.0 & \cellcolor[rgb]{0.635,0.839,0.686}14.0 & \cellcolor[rgb]{0.878,0.952,0.865}9.0 & \cellcolor[rgb]{0.754,0.897,0.753}20.1 \\
+ Ours & \cellcolor[rgb]{0.500,0.633,0.553}54.0 & \cellcolor[rgb]{0.635,0.839,0.686}13.0 & \cellcolor[rgb]{0.635,0.839,0.686}32.0 & \cellcolor[rgb]{0.500,0.633,0.553}33.7 & \cellcolor[rgb]{0.500,0.633,0.553}45.2 & \cellcolor[rgb]{0.500,0.633,0.553}47.0 & \cellcolor[rgb]{0.878,0.952,0.865}17.0 & \cellcolor[rgb]{0.500,0.633,0.553}22.0 & \cellcolor[rgb]{0.500,0.633,0.553}22.0 & \cellcolor[rgb]{0.500,0.633,0.553}31.8 \\
\bottomrule
\end{tabular}
}
\caption{Performance heatmap by methods and datasets. All tasks are evaluated on the HARD set with Pass@1 Accuracy (\%) using GPT-4o base model.}
  \label{tab:results_mas_gpt-4o}
\end{table*}

\begin{table*}[htbp]
  \centering
  \resizebox{\textwidth}{!}{
\begin{tabular}{l|c|c|c|c|c|c|c|c|c|c}
\toprule
Method & MedQA & PubMedQA & MedMCQA & MedBullets & MMLU & MMLU-Pro & MedExQA & MedXpert-R & MedXpert-U & Average \\
\midrule
DeepSeek-R1 & \cellcolor[rgb]{0.932,0.974,0.921}38.0 & \cellcolor[rgb]{0.984,0.994,0.980}11.0 & \cellcolor[rgb]{0.932,0.974,0.921}28.0 & \cellcolor[rgb]{0.932,0.974,0.921}36.0 & \cellcolor[rgb]{0.932,0.974,0.921}32.9 & \cellcolor[rgb]{0.829,0.931,0.817}36.0 & \cellcolor[rgb]{0.984,0.994,0.980}20.0 & \cellcolor[rgb]{0.829,0.931,0.817}20.0 & \cellcolor[rgb]{0.829,0.931,0.817}23.0 & \cellcolor[rgb]{0.829,0.931,0.817}27.2 \\
+ few-shot & \cellcolor[rgb]{0.984,0.994,0.980}27.0 & \cellcolor[rgb]{0.932,0.974,0.921}12.0 & \cellcolor[rgb]{0.576,0.781,0.642}32.0 & \cellcolor[rgb]{0.984,0.994,0.980}33.7 & \cellcolor[rgb]{0.932,0.974,0.921}35.6 & \cellcolor[rgb]{0.576,0.781,0.642}41.0 & \cellcolor[rgb]{0.500,0.633,0.553}27.0 & \cellcolor[rgb]{0.932,0.974,0.921}11.0 & \cellcolor[rgb]{0.932,0.974,0.921}9.0 & \cellcolor[rgb]{0.932,0.974,0.921}25.4 \\
+ RAG & \cellcolor[rgb]{0.701,0.871,0.718}49.0 & \cellcolor[rgb]{0.576,0.781,0.642}20.0 & \cellcolor[rgb]{0.701,0.871,0.718}31.0 & \cellcolor[rgb]{0.576,0.781,0.642}43.8 & \cellcolor[rgb]{0.505,0.718,0.590}53.8 & \cellcolor[rgb]{0.505,0.718,0.590}42.0 & \cellcolor[rgb]{0.701,0.871,0.718}25.0 & \cellcolor[rgb]{0.505,0.718,0.590}28.0 & \cellcolor[rgb]{0.576,0.781,0.642}26.0 & \cellcolor[rgb]{0.505,0.718,0.590}35.4 \\
+ CoT & \cellcolor[rgb]{0.829,0.931,0.817}47.0 & \cellcolor[rgb]{0.932,0.974,0.921}12.0 & \cellcolor[rgb]{0.701,0.871,0.718}31.0 & \cellcolor[rgb]{0.829,0.931,0.817}39.3 & \cellcolor[rgb]{0.829,0.931,0.817}38.4 & \cellcolor[rgb]{0.829,0.931,0.817}35.0 & \cellcolor[rgb]{0.932,0.974,0.921}22.0 & \cellcolor[rgb]{0.576,0.781,0.642}27.0 & \cellcolor[rgb]{0.505,0.718,0.590}27.0 & \cellcolor[rgb]{0.829,0.931,0.817}31.0 \\
+ CoT-SC & \cellcolor[rgb]{0.505,0.718,0.590}52.0 & \cellcolor[rgb]{0.829,0.931,0.817}14.0 & \cellcolor[rgb]{0.576,0.781,0.642}32.0 & \cellcolor[rgb]{0.576,0.781,0.642}43.8 & \cellcolor[rgb]{0.576,0.781,0.642}45.2 & \cellcolor[rgb]{0.701,0.871,0.718}38.0 & \cellcolor[rgb]{0.829,0.931,0.817}24.0 & \cellcolor[rgb]{0.829,0.931,0.817}17.0 & \cellcolor[rgb]{0.576,0.781,0.642}26.0 & \cellcolor[rgb]{0.701,0.871,0.718}32.4 \\
+ Multi-Persona & \cellcolor[rgb]{0.505,0.718,0.590}52.0 & \cellcolor[rgb]{0.701,0.871,0.718}18.0 & \cellcolor[rgb]{0.500,0.633,0.553}37.0 & \cellcolor[rgb]{0.701,0.871,0.718}42.7 & \cellcolor[rgb]{0.829,0.931,0.817}42.5 & \cellcolor[rgb]{0.701,0.871,0.718}38.0 & \cellcolor[rgb]{0.576,0.781,0.642}26.0 & \cellcolor[rgb]{0.701,0.871,0.718}23.0 & \cellcolor[rgb]{0.576,0.781,0.642}26.0 & \cellcolor[rgb]{0.576,0.781,0.642}33.9 \\
+ Self-Refine & \cellcolor[rgb]{0.932,0.974,0.921}33.0 & \cellcolor[rgb]{0.701,0.871,0.718}17.0 & \cellcolor[rgb]{0.829,0.931,0.817}30.0 & \cellcolor[rgb]{0.932,0.974,0.921}34.8 & \cellcolor[rgb]{0.984,0.994,0.980}27.4 & \cellcolor[rgb]{0.984,0.994,0.980}22.0 & \cellcolor[rgb]{0.829,0.931,0.817}24.0 & \cellcolor[rgb]{0.932,0.974,0.921}12.0 & \cellcolor[rgb]{0.932,0.974,0.921}13.0 & \cellcolor[rgb]{0.984,0.994,0.980}23.7 \\
+ MedAgents & \cellcolor[rgb]{0.576,0.781,0.642}48.0 & \cellcolor[rgb]{0.576,0.781,0.642}21.0 & \cellcolor[rgb]{0.984,0.994,0.980}22.0 & \cellcolor[rgb]{0.576,0.781,0.642}44.9 & \cellcolor[rgb]{0.701,0.871,0.718}43.8 & \cellcolor[rgb]{0.829,0.931,0.817}35.0 & \cellcolor[rgb]{0.500,0.633,0.553}27.0 & \cellcolor[rgb]{0.576,0.781,0.642}22.0 & \cellcolor[rgb]{0.701,0.871,0.718}25.0 & \cellcolor[rgb]{0.701,0.871,0.718}32.1 \\
+ MedPrompt & \cellcolor[rgb]{0.576,0.781,0.642}46.0 & \cellcolor[rgb]{0.829,0.931,0.817}14.0 & \cellcolor[rgb]{0.829,0.931,0.817}30.0 & \cellcolor[rgb]{0.829,0.931,0.817}38.2 & \cellcolor[rgb]{0.576,0.781,0.642}45.2 & \cellcolor[rgb]{0.932,0.974,0.921}27.0 & \cellcolor[rgb]{0.829,0.931,0.817}24.0 & \cellcolor[rgb]{0.984,0.994,0.980}8.0 & \cellcolor[rgb]{0.984,0.994,0.980}7.0 & \cellcolor[rgb]{0.932,0.974,0.921}26.6 \\
+ Ours & \cellcolor[rgb]{0.500,0.633,0.553}55.0 & \cellcolor[rgb]{0.500,0.633,0.553}28.0 & \cellcolor[rgb]{0.505,0.718,0.590}35.0 & \cellcolor[rgb]{0.500,0.633,0.553}52.8 & \cellcolor[rgb]{0.500,0.633,0.553}61.6 & \cellcolor[rgb]{0.500,0.633,0.553}53.0 & \cellcolor[rgb]{0.701,0.871,0.718}26.0 & \cellcolor[rgb]{0.500,0.633,0.553}36.0 & \cellcolor[rgb]{0.500,0.633,0.553}49.0 & \cellcolor[rgb]{0.500,0.633,0.553}44.0 \\
\bottomrule
\end{tabular}
}
\caption{Performance heatmap by methods and datasets. All tasks are evaluated on the HARD set with Pass@1 Accuracy (\%) using DeepSeek-R1 base model.}
\label{tab:results_mas_DeepSeek-R1}
\end{table*}

\section{Experiments}
\label{sec:experiments}

\subsection{Experimental Setting}
\textbf{Retrieval Dataset.}
We use the MedReason dataset~\cite{wu2025medreason} as our external retrieval corpus. It contains 32,682 medical QA pairs with clinically validated, step-by-step explanations generated via a knowledge graph–guided pipeline based on PrimeKG~\cite{chandak2023building}. Unlike general CoT datasets, MedReason ensures factual correctness by filtering out chains that do not lead to the correct answer. We treat it as a structured knowledge base for retrieving semantically or logically relevant examples at inference. Dataset construction details and examples are provided in Appendix~\ref{app_refenrence_medreason}.

\textbf{Test Datasets.} 
We evaluate TAGS on a curated benchmark of nine medical QA datasets selected from the MEDAGENTSBENCH framework~\cite{tang2025medagentsbench}, designed to assess complex clinical reasoning. The benchmark includes challenging subsets from:
\textbf{MedQA}~\cite{jin2021medqa}, a multilingual board-exam dataset (e.g., USMLE);
\textbf{PubMedQA}~\cite{jin2019pubmedqa}, derived from biomedical literature with yes/no/maybe answers;
\textbf{MedMCQA}~\cite{pal2022medmcqa}, covering 21 medical subjects from Indian medical exams;
\textbf{MedBullets}~\cite{chen2024medbullet}, featuring long-context clinical questions;
\textbf{MedExQA}~\cite{kim2024medexqa}, emphasizing explainable QA across five specialties;
\textbf{MedXpertQA}~\cite{zuo2025medxpertqa}, with subsets targeting reasoning and understanding;
\textbf{MMLU}~\cite{hendrycks2020mmlu} and \textbf{MMLU-Pro} \cite{wang2024mmlu}, general benchmarks with medical subfields.

To better reflect real-world difficulty, we follow the hard subset construction pipeline proposed by MEDAGENTSBENCH. Questions are selected based on model failure rates (<50\% accuracy across a set of strong models), medical topic coverage, and reasoning depth. Specifically, we include 100 hard questions each from MedQA, PubMedQA, MedMCQA, MedExQA, and MMLU-Pro; 100 from each MedXpertQA subset (Reasoning and Understanding); 89 from MedBullets; and 73 from MMLU. This results in a total of 862 expert-verified instances designed to stress-test the reasoning capabilities of large language models.

\begin{table*}[t]
  \centering
  \resizebox{\textwidth}{!}{
\begin{tabular}{l|c|c|c|c|c|c|c|c|c|c}
\toprule
Method & MedQA & PubMedQA & MedMCQA & MedBullets & MMLU & MMLU-Pro & MedExQA & MedXpert-R & MedXpert-U & Average \\
\midrule
Qwen2.5-7B & \cellcolor[rgb]{0.984,0.994,0.980}16.0 & \cellcolor[rgb]{0.919,0.968,0.907}16.0 & \cellcolor[rgb]{0.857,0.943,0.844}24.0 & \cellcolor[rgb]{0.984,0.994,0.980}4.5 & \cellcolor[rgb]{0.953,0.982,0.944}13.7 & \cellcolor[rgb]{0.500,0.704,0.582}26.0 & \cellcolor[rgb]{0.984,0.994,0.980}9.0 & \cellcolor[rgb]{0.878,0.952,0.865}10.0 & \cellcolor[rgb]{0.917,0.968,0.906}8.0 & \cellcolor[rgb]{0.984,0.994,0.980}14.1 \\
\midrule
Llama-3-8B & \cellcolor[rgb]{0.936,0.975,0.925}18.0 & \cellcolor[rgb]{0.984,0.994,0.980}13.0 & \cellcolor[rgb]{0.899,0.960,0.886}23.0 & \cellcolor[rgb]{0.597,0.803,0.658}16.9 & \cellcolor[rgb]{0.984,0.994,0.980}11.0 & \cellcolor[rgb]{0.552,0.758,0.623}23.0 & \cellcolor[rgb]{0.936,0.975,0.925}11.0 & \cellcolor[rgb]{0.878,0.952,0.865}10.0 & \cellcolor[rgb]{0.974,0.990,0.969}4.0 & \cellcolor[rgb]{0.977,0.991,0.971}14.4 \\
\midrule
HuatuoGPT-o1-7B & \cellcolor[rgb]{0.754,0.897,0.753}22.0 & \cellcolor[rgb]{0.726,0.884,0.731}21.0 & \cellcolor[rgb]{0.754,0.897,0.753}26.0 & \cellcolor[rgb]{0.770,0.904,0.767}12.4 & \cellcolor[rgb]{0.953,0.982,0.944}13.7 & \cellcolor[rgb]{0.500,0.633,0.553}29.0 & \cellcolor[rgb]{0.984,0.994,0.980}9.0 & \cellcolor[rgb]{0.822,0.928,0.810}11.0 & \cellcolor[rgb]{0.936,0.975,0.925}7.0 & \cellcolor[rgb]{0.876,0.951,0.862}16.8 \\
\midrule
HuatuoGPT-o1-8B & \cellcolor[rgb]{0.500,0.633,0.553}29.0 & \cellcolor[rgb]{0.770,0.904,0.767}20.0 & \cellcolor[rgb]{0.500,0.633,0.553}33.0 & \cellcolor[rgb]{0.500,0.714,0.586}20.2 & \cellcolor[rgb]{0.768,0.903,0.764}21.9 & \cellcolor[rgb]{0.682,0.862,0.709}17.0 & \cellcolor[rgb]{0.595,0.801,0.656}18.0 & \cellcolor[rgb]{0.530,0.740,0.608}16.0 & \cellcolor[rgb]{0.936,0.975,0.925}7.0 & \cellcolor[rgb]{0.629,0.836,0.683}20.2 \\
\midrule
MedLlama-3-8B-v1.0 & \cellcolor[rgb]{0.635,0.839,0.686}24.0 & \cellcolor[rgb]{0.770,0.904,0.767}20.0 & \cellcolor[rgb]{0.936,0.975,0.925}22.0 & \cellcolor[rgb]{0.679,0.860,0.708}14.6 & \cellcolor[rgb]{0.904,0.963,0.892}16.4 & \cellcolor[rgb]{0.819,0.927,0.808}12.0 & \cellcolor[rgb]{0.899,0.960,0.886}12.0 & \cellcolor[rgb]{0.822,0.928,0.810}11.0 & \cellcolor[rgb]{0.857,0.943,0.844}11.0 & \cellcolor[rgb]{0.921,0.969,0.909}15.9 \\
\midrule
MedLlama-3-8B-v2.0 & \cellcolor[rgb]{0.500,0.681,0.573}28.0 & \cellcolor[rgb]{0.567,0.771,0.634}25.0 & \cellcolor[rgb]{0.936,0.975,0.925}22.0 & \cellcolor[rgb]{0.500,0.633,0.553}22.5 & \cellcolor[rgb]{0.500,0.704,0.582}32.9 & \cellcolor[rgb]{0.819,0.927,0.808}12.0 & \cellcolor[rgb]{0.500,0.633,0.553}22.0 & \cellcolor[rgb]{0.923,0.970,0.911}9.0 & \cellcolor[rgb]{0.917,0.968,0.906}8.0 & \cellcolor[rgb]{0.629,0.836,0.683}20.2 \\
\midrule
OpenBioLLM-8B & \cellcolor[rgb]{0.899,0.960,0.886}19.0 & \cellcolor[rgb]{0.500,0.633,0.553}29.0 & \cellcolor[rgb]{0.984,0.994,0.980}20.0 & \cellcolor[rgb]{0.535,0.743,0.611}19.1 & \cellcolor[rgb]{0.768,0.903,0.764}21.9 & \cellcolor[rgb]{0.984,0.994,0.980}2.0 & \cellcolor[rgb]{0.635,0.839,0.686}17.0 & \cellcolor[rgb]{0.984,0.994,0.980}7.0 & \cellcolor[rgb]{0.984,0.994,0.980}3.0 & \cellcolor[rgb]{0.950,0.981,0.940}15.3 \\
\midrule
Qwen2.5-7B + Ours & \cellcolor[rgb]{0.500,0.681,0.573}28.0 & \cellcolor[rgb]{0.567,0.771,0.634}25.0 & \cellcolor[rgb]{0.857,0.943,0.844}24.0 & \cellcolor[rgb]{0.679,0.860,0.708}14.6 & \cellcolor[rgb]{0.500,0.633,0.553}35.6 & \cellcolor[rgb]{0.511,0.723,0.594}25.0 & \cellcolor[rgb]{0.698,0.870,0.717}16.0 & \cellcolor[rgb]{0.500,0.633,0.553}18.0 & \cellcolor[rgb]{0.500,0.633,0.553}29.0 & \cellcolor[rgb]{0.500,0.633,0.553}23.9 \\
\midrule
\bottomrule
\end{tabular}}
\caption{Comparison with fine-tuned medical LLMs on nine MedQA benchmarks.}
\label{tab:results_medllm}
\end{table*}

\textbf{Baselines.} 
We first compare our method against several widely adopted prompting and reasoning strategies that do not involve model updates:
\textbf{(1) CoT (Chain-of-Thought)}\cite{wei2022chain}: A prompting technique that guides the model to articulate intermediate reasoning steps before producing a final answer.
\textbf{(2) CoT-SC (Chain-of-Thought with Self-Consistency)}\cite{wang2022self}: An extension of CoT that generates multiple reasoning paths and selects the most consistent answer via majority voting.
\textbf{(3) Multi-Persona}\cite{wang2023multipersona}: A method that simulates multiple expert personas to collaboratively reason through clinical questions.
\textbf{(4) Self-Refine}\cite{madaan2024selfrefine}: A self-improvement framework in which the model iteratively refines its own responses across multiple reasoning stages.
\textbf{(5) MedAgents}\cite{tang2023medagents}: A domain-specific multi-agent framework that employs multiple specialist agents for collaborative clinical reasoning.
\textbf{(6) MDAgents}\cite{kim2024mdagents}: A lightweight variant of MedAgents that combines minimal agent collaboration with retrieval augmentation to improve reasoning.
\textbf{(7) MedPrompt}~\cite{chen2024medprompt}: A retrieval-augmented prompting strategy that integrates semantically similar historical cases to enhance clinical inference.

We additionally report results under a \textbf{few-shot} baseline, where five training examples from the target dataset are retrieved and used as in-context demonstrations for single-pass inference. We also include a \textbf{RAG} baseline, which retrieves the top-$K$ most semantically similar questions with accompanying CoTs from the MedReason dataset and feeds them directly to the model. This RAG setting shares the same retrieval setup but excludes agent collaboration and verification, highlighting the value of structured reasoning.

We further evaluate our method against several strong open-source foundation models and their medically adapted variants:
\textbf{(1) Qwen2.5-7B}~\cite{yang2024qwen2}: A 7B general-purpose model instruction-tuned for diverse tasks, evaluated both with and without CoT prompting.
\textbf{(2) LLaMA-3-8B}~\cite{grattafiori2024llama}: Meta’s latest 8B instruction-tuned model with improved reasoning capabilities.
\textbf{(3) HuatuoGPT-o1-7B}~\cite{chen2024huatuogpt}: A 7B model fine-tuned for complex medical reasoning via reinforcement learning.
\textbf{(4) HuatuoGPT-o1-8B}~\cite{chen2024huatuogpt}: An enhanced 8B version of HuatuoGPT, optimized for clinical inference tasks.
\textbf{(5) MedLLaMA-3-8B-v1.0}~\cite{qiu2024towards}: A medical-adapted variant of LLaMA-3 trained on biomedical corpora.
\textbf{(6) MedLLaMA-3-8B-v2.0}: An updated release with improved performance on expert-level medical benchmarks.
\textbf{(7) OpenBioLLM-8B}~\cite{pal2024openbiollms}: An open-source 8B biomedical language model fine-tuned for healthcare and life sciences applications.

\textbf{Evaluation Metrics} Following~\cite{tang2023medagents}, we report \textbf{Pass@1 Accuracy} as the evaluation metric, which measures whether the model’s first generated answer exactly matches the ground-truth answer.

\textbf{Reproducibility}. All experiments were conducted using Python 3.10 and PyTorch 2.4.0 on four NVIDIA H100 GPUs, each with 80 GB of memory. For proprietary LLM baselines such as GPT-4o and DeepSeek-R1, we accessed the models through their official APIs and ensured consistent use of the same model version across all runs. For open-source models, we directly loaded checkpoint weights from their respective official Hugging Face repositories to ensure reproducibility and transparency.

\subsection{Compared with Prompting and MAS}
\label{subsec:comparison_mas}

We evaluate the effectiveness of \textsc{TAGS} by comparing it with a diverse set of prompting-based and multi-agent reasoning baselines across nine challenging MedQA benchmarks. Tables~\ref{tab:results_mas_gpt-4o} and~\ref{tab:results_mas_DeepSeek-R1} summarize the results in terms of \textbf{Pass@1 Accuracy}, evaluated on the HARD split using two foundational LLMs: GPT-4o and DeepSeek-R1.
Under the GPT-4o setting, \textsc{TAGS} achieves the highest average accuracy of 31.8\%, outperforming all baselines including Self-Refine (26.9\%), MedAgents (20.0\%), and CoT-SC (25.3\%). The most notable improvements appear on MedQA (+9.0 over Multi-Persona), MMLU (+8.2 over Multi-Persona), and MedXpert-R (+3.0 over RAG), highlighting the impact of verifier-guided aggregation and structured multi-agent reasoning. \textsc{TAGS} also surpasses standard few-shot and RAG baselines by margins of +10.1 and +8.7 respectively.
With the DeepSeek-R1 base model, \textsc{TAGS} achieves an average accuracy of 44.0\%, outperforming CoT-SC (32.4\%), Multi-Persona (33.9\%), and MedAgents (32.1\%). Notably, MDAgents consistently failed under this setting due to format inconsistencies. \textsc{TAGS} also surpasses the few-shot and RAG baselines by margins of +18.6 and +8.5, respectively, demonstrating the scalability of our framework across both general and domain-specific tasks.

\subsection{Compared with Fine-Tuned LLMs}
\label{subsec:comparison_medllm}
To further contextualize the performance of our \textsc{TAGS} framework, we evaluate its effectiveness when integrated with the Qwen2.5-7B base model and compare its performance against a series of prominent open-source and medically fine-tuned large language models across the same nine challenging MedQA datasets. The results of this comparison are presented in Table~\ref{tab:results_medllm}. As shown in the table, our \textsc{TAGS} framework substantially boosts the zero-shot question answering capability of the base Qwen2.5-7B model, improving its average accuracy from 14.1\% to 23.9\%.
In particular, \textsc{TAGS} demonstrates robust performance gains on difficult benchmarks such as MedQA (+12.0 percentage points), MMLU (+21.9 percentage points), and MedXpert-U (+21.0 percentage points). Notably, our inference-only strategy even outperforms several models that have been fine-tuned with domain-specific medical corpora or expert feedback, such as MedLLaMA-3-8B and the HuatuoGPT-o1 variants, on the majority of the evaluated datasets. These results strongly highlight the significant potential of structured retrieval and multi-agent reasoning, combined with uncertainty-aware verification, to effectively close the performance gap with models requiring extensive fine-tuning, while retaining the inherent flexibility and adaptability of a zero-shot approach.

\subsection{Ablation Study}
\label{ablation}

We conduct an ablation study on the Qwen2.5-7B model to assess the impact of each module in the \textsc{TAGS} framework, focusing on two datasets: MMLU and MedXpert-U. 
As shown in Table~\ref{tab:ablation_results}, the baseline Qwen2.5-7B model without any module achieves 13.7\% and 8.0\% accuracy on MMLU and MedXpert-U, respectively. Incorporating retrieval augmentation (RAG) improves the scores to 20.5\% and 10.0\%, highlighting the utility of external knowledge retrieval.  
Adding the generalist (G) and specialist (S) agents with a majority voting mechanism raises MMLU accuracy to 30.1\%, while MedXpert-U remains at 10.0\%, indicating limited impact on domain-specific tasks.
Integrating hierarchical retrieval augmentation (HRA) results in notable gains, with 34.2\% for the generalist and 32.9\% for the specialist, indicating the value of reasoning-guided retrieval.
The complete \textsc{TAGS} framework, including uncertainty-aware aggregation (UAAA), achieves the best performance of 35.6\% and 29.0\%, confirming the synergistic impact of structured retrieval, dual-agent reasoning, and verification.  

To assess whether our framework relies on retrieving semantically closest examples, we introduce \textbf{RAG-w/o-topk}, which explicitly excludes the top-10 most similar questions during retrieval. Despite this restriction, performance only marginally decreases, demonstrating that our model benefits primarily from exposure to valid reasoning patterns rather than from copying specific answers. We further introduce \textbf{$\mathcal{S}$-w-3rd}, a variant that assigns the 3rd most relevant specialist instead of the top-ranked one. This setting simulates scenarios where the domain classifier misidentifies the optimal expert, which may occur in real-world deployments. The performance drops only slightly under this perturbation, suggesting that TAGS does not strongly depend on perfect specialist selection. Even suboptimal specialists can provide useful guidance, highlighting the robustness of our framework.

\begin{table}[t]
  \centering
  \resizebox{0.5\textwidth}{!}{
    \begin{tabular}{ccccc|c|c}
    \toprule
    \multicolumn{1}{c}{RAG} & 
\multicolumn{1}{c}{$\mathcal{G}$} & 
\multicolumn{1}{c}{$\mathcal{S}$} & 
\multicolumn{1}{c}{HRA} & 
\multicolumn{1}{c|}{UAAA} & 
\multicolumn{1}{c|}{MMLU} & 
\multicolumn{1}{c}{MedXpert-U} \\
    \hline
          &       &       &       &       & 13.7  & 8.0  \\
    \hline
     \cmark    &       &       &       &       & 20.5  & 10.0  \\
    \hline
    \cmark     & \cmark     & \cmark     &       &       & 30.1  & 10.0  \\
    \hline
    \cmark     & \cmark     &       & \cmark     & \cmark     & 34.2  & 16.0  \\
    \hline
    \cmark     &       & \cmark     & \cmark     & \cmark     & 32.9  & 18.0  \\
    \hline
    \cmark     & \cmark     & \cmark     & \cmark     &       & 31.5  & 22.0  \\
    
    \hline
    \cmark     & \cmark     & \cmark     & \cmark     & \cmark     & \cellcolor[rgb]{0.500,0.681,0.573}35.6  & \cellcolor[rgb]{0.500,0.681,0.573}29.0 \\ \hline 
    w/o-top10    & \cmark     & \cmark     & \cmark     &  \cmark     & 37.0  & 24.0  \\
    \cmark    & \cmark     & w-3rd     & \cmark     &  \cmark     & 34.2  & 23.0  \\
    \bottomrule
    \end{tabular}    
    }
   \caption{Ablation Study on \textsc{TAGS} using Qwen2.5-7B. 
RAG: Retrieval-Augmented Generation; 
$\mathcal{G}$: Generalist Agent; 
$\mathcal{S}$: Specialist Agent; 
HRA: Hierarchical Retrieval Augmentation; 
UAAA: Uncertainty-Aware Answer Aggregation.}
  \label{tab:ablation_results}%
\end{table}%

\subsection{Hyperparameter Analysis}
\label{subsec:hyperparam}

\begin{figure}
    \centering
    \includegraphics[width=\linewidth]{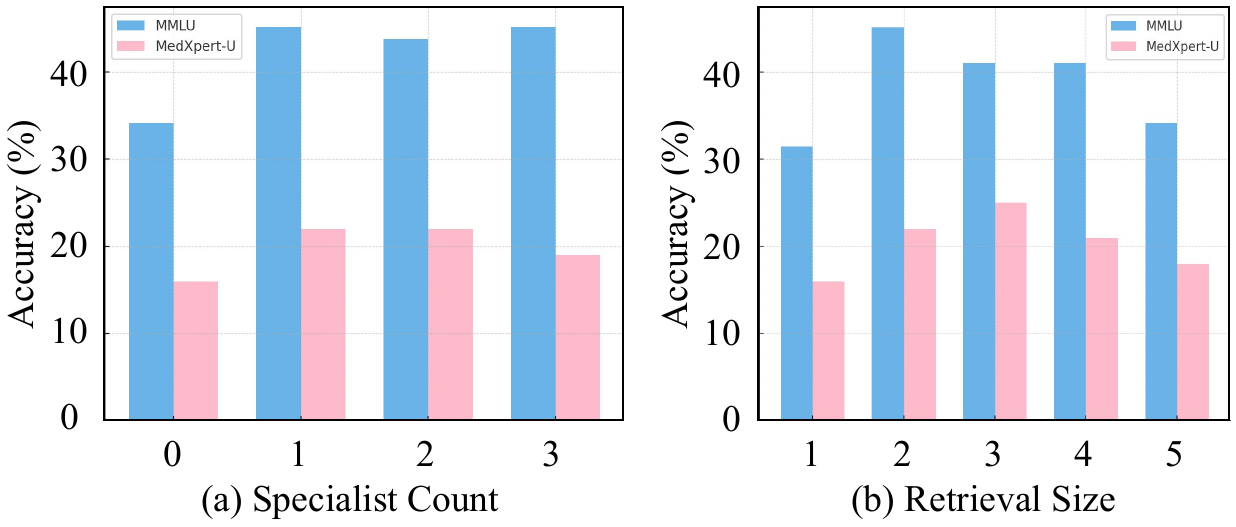}
    \caption{Hyper-parameter sensitivity analysis of specialist count and retrieval size in relation to accuracy.}
    \label{fig_hypa_2}
\end{figure}

Figure~\ref{fig_hypa_2} shows TAGS’ sensitivity to two key hyperparameters: the number of specialist agents and the retrieval size \(K\).
In Figure~\ref{fig_hypa_2}(a), adding one specialist to the generalist improves accuracy from 34.2\% to 45.2\% on MMLU and from 16.0\% to 22.0\% on MedXpert-U. However, adding more specialists brings limited or no further gains, likely due to redundancy or conflicts in reasoning paths.
Figure~\ref{fig_hypa_2}(b) shows that accuracy peaks at \(K=2\) and declines with larger \(K\), as additional exemplars may introduce noise or irrelevant content that misleads the model.
These results support our choice of using one specialist and \(K=2\) as the default configuration, balancing diversity and robustness.

\subsection{Inference Efficiency}
On the MedQA dataset with GPT-4o, \textsc{TAGS} takes 72 seconds per question on average, which is longer than CoT-SC (27.7s) but shorter than Multi-Persona (109.6s). Although slower than simple prompting, \textsc{TAGS} achieves substantially higher accuracy. Both reasoning and verification are parallelizable, enabling efficient deployment in real-world clinical settings. This moderate inference cost represents a favorable trade-off for improved robustness and reliability. Additionally, these stages can be parallelized across GPU streams or executed via asynchronous API calls to further speed up inference.


\section{Conclusion}
We presented TAGS, a parameter-efficient and test-time-only framework designed to enhance reliability in medical question answering without model fine-tuning. TAGS introduces a structured reasoning paradigm through generalist–specialist reasoning collaboration, which combines the breadth of a generalist with the depth of a specialist to generate complementary inference paths. This collaboration is guided by hierarchical retrieval augmentation, which retrieves exemplars at both semantic and rationale levels to enrich reasoning diversity, and finalized by uncertainty-aware answer aggregation to select robust answers.
Extensive experiments on nine challenging MedQA benchmarks, spanning general-purpose and fine-tuned LLMs, demonstrate TAGS’ consistent superiority over prompting-based, retrieval-augmented, and multi-agent baselines. Notably, our method delivers substantial improvements even for compact 7B models, highlighting its adaptability across model scales.
TAGS offers a practical, inference-only alternative for trustworthy medical AI and opens promising directions for adaptive retrieval, dynamic agent collaboration, and scaling to multimodal or real-world clinical QA workflows.

\section*{Limitations}

While TAGS offers a robust, inference-only approach to medical QA, it carries several limitations. First, it depends heavily on the coverage and quality of the external retrieval corpus: gaps or biases in the QA database may lead to missing or misleading exemplars, particularly for rare diseases or newly emerging clinical scenarios. Second, the Reasoning Consistency Evaluator (RCE) is itself a zero-shot LLM prompt and may inherit the same hallucination tendencies or biases as the generator agents, potentially mis-scoring perfectly valid but unconventional reasoning chains. Third, the two-round retrieval and dual-agent design, while effective, substantially increases inference latency and API cost compared to single-pass prompting; this may limit real-time deployment in resource-constrained clinical settings.

Additionally, our current evaluation focuses solely on answer accuracy (Pass@1), without assessing the interpretability or faithfulness of reasoning paths. Future work may benefit from human evaluation or rationale consistency metrics to further assess clinical applicability.

Moreover, our specialty inference component can occasionally misclassify the most relevant domain, which, although gracefully handled, may still introduce suboptimal reasoning contexts.
Finally, our evaluation is confined to English-language, multiple-choice benchmarks and does not cover open-ended clinical dialogs, multimodal data (e.g., images, lab reports), or non-English patient populations. Addressing these limitations will require enriching and updating the retrieval corpus, developing more calibrated or human-in-the-loop verifier mechanisms, optimizing retrieval budgets and round counts, and extending evaluation to diverse, real-world clinical workflows.

\bibliography{main}

\newpage

\appendix
\tcbset{
  colback=gray!3,      
  colframe=gray!50,    
  boxrule=0.4pt,       
  arc=2mm,             
  left=2pt,            
  right=2pt,           
  top=2pt,             
  bottom=2pt,          
  width=0.5\textwidth,    
  boxsep=2pt           
}

\lstset{
    basicstyle=\footnotesize\ttfamily,
    breaklines=true,         
    breakatwhitespace=true,  
    frame=single,            
    backgroundcolor=\color{gray!10},
    linewidth=\textwidth     
}

\section{Ethics Statement}
This work relies solely on publicly available medical question answering datasets, including MedQA, PubMedQA, MedMCQA, and others curated in the MEDAGENTSBENCH framework~\cite{tang2025medagentsbench}. These datasets are de-identified and collected from open educational or biomedical sources such as medical board exams and peer-reviewed literature. No private health records or patient-identifiable information were used.

Our proposed framework operates entirely at test time and does not require any model fine-tuning or user data collection. All evaluations are conducted offline on benchmark datasets, and no deployment in real clinical settings has been performed.

While our method is designed to improve the robustness and reliability of medical LLMs, it is not intended for use in high-stakes clinical decision-making without appropriate human oversight. We emphasize that the generated answers should not be interpreted as medical advice. Future work may involve incorporating human-in-the-loop mechanisms and broader impact assessments before real-world deployment.

\section{Generalist–Specialist System Roles and Prompt Templates}\label{Generalist–Specialist System Roles and Prompt}

\subsection{System Prompt for Specialist and Generalist}

To ensure consistency and clarity across different model roles, we define structured system prompts tailored to each classifier in our multi-agent framework. These prompts specify role-specific reasoning strategies and output formats, enabling the models to adopt appropriate clinical reasoning behaviors under zero-shot test-time conditions.

The system prompt for the specialist categorization agent is presented in Table~\ref{app_specialist_categorization}, while the diagnostic prompt for the specialist agent is shown in Table~\ref{app_tab:specialist_prompt}. The prompt for the generalist agent is provided in Table~\ref{app_tab:generalist_prompt}.

\begin{table}[htbp]
\centering
\begin{tcolorbox}[enhanced, sharp corners, colback=gray!8, colframe=black!20, boxrule=0.5pt, width=0.5\textwidth, title={\textbf{Specialist Categorization — System Prompt}}]
\footnotesize
You are a senior medical expert tasked with classifying clinical multiple-choice problems into the most relevant areas of medical science.\\[4pt]

Your role is strictly to determine and output the classification.\\

\textbf{Important:} \emph{Do not provide any explanation, reasoning, or commentary. Only output the final classification strictly following the format.} \\

\end{tcolorbox}
\caption{System prompt for the specialist categorization.}
\label{app_specialist_categorization}
\end{table}

\begin{table}[htbp]
\centering
\begin{tcolorbox}[enhanced, sharp corners, colback=gray!8, colframe=black!20, boxrule=0.5pt, width=0.5\textwidth, title={\textbf{Specialist Agent — System Prompt}}]
\footnotesize
You are an experienced specialist in \{\texttt{domain}\}. Your role is to carefully
analyze clinical multiple-choice questions from the standpoint of a
\{\texttt{domain.lower()}\} expert. You should reason by focusing on the interpretation
of symptoms, underlying pathophysiology, and domain-specific diagnostic principles.\\[4pt]

First, review the provided reference examples and understand their reasoning patterns.\\
Then, based on your specialist knowledge, perform \emph{structured, step-by-step} reasoning
for the new question.\\[6pt]

\textbf{Required output format}\\
\texttt{Thought: [your detailed step-by-step reasoning]}\\
\texttt{Answer: [one of A, B, C, …]}
\end{tcolorbox}
\caption{System prompt for the specialist agent.}
\label{app_tab:specialist_prompt}
\end{table}

\begin{table}[htbp]
\centering
\begin{tcolorbox}[enhanced, sharp corners, colback=gray!8, colframe=black!20, boxrule=0.5pt, width=0.5\textwidth, title={\textbf{Generalist Agent — System Prompt}}]
\footnotesize
You are a general practitioner trained to manage a wide range of clinical conditions.  
Your task is to evaluate clinical multiple-choice questions using broad, cross-disciplinary medical knowledge. Focus on extracting key clinical findings, ruling out unlikely diagnoses, and applying general reasoning principles.  \\[4pt]

First, analyze the reference examples to understand their diagnostic thought process.  \\
Then, produce a \emph{step-by-step} analysis for the new question.   \\[6pt]

\textbf{Required output format}   \\
\texttt{Thought: [your detailed step-by-step reasoning]}  \\
\texttt{Answer: [one of A, B, C, …]} \\

\end{tcolorbox}
\caption{System prompt for the generalist agent.}
\label{app_tab:generalist_prompt}
\end{table}

\subsection{Prompt Organization and Structure}

To ensure faithful and consistent model behavior across different roles and stages of inference, we design modular and task-specific prompt templates. These templates guide the models in both few-shot reasoning and auxiliary classification tasks.

Specifically, the specialist classification prompt (Table~\ref{tab:specialist_classification_prompt}) is used to determine the relevant subfields of medicine needed to solve a given question, serving as a basis for downstream role assignment and retrieval. Meanwhile, the few-shot prompt template (Table~\ref{tab:fewshot_prompt}) provides structured instructions and reference examples to facilitate reasoning transfer for clinical question answering. 

\begin{table}[htbp]
\centering
\begin{tcolorbox}[
  enhanced,
  sharp corners,
  colback=gray!8,           
  colframe=black!20,        
  boxrule=0.5pt,
  width=0.5\textwidth,
  title={\textbf{Specialist Classifier — Prompt}}
]
\footnotesize
\textbf{Task Instructions}
\begin{itemize}[leftmargin=*, noitemsep]
  \item Carefully analyze the following medical question:
    \begin{verbatim}
'''{question}'''
    \end{verbatim}

  \item The corresponding options are:
    \begin{verbatim}
'''{options}'''
    \end{verbatim}

  \item Based on both the question and the options, determine the top \textbf{3 most relevant subfields of medicine} that are required to solve this question.

  \item You must only output in the exact format:
    \begin{verbatim}
Medical Field: Field1 | Field2 | Field3
    \end{verbatim}
\end{itemize}
\end{tcolorbox}
\caption{Prompt used for classifying medical questions into relevant specialist subfields.}
\label{tab:specialist_classification_prompt}
\end{table}

\begin{table}[t]
\centering
\begin{tcolorbox}[
  enhanced,
  sharp corners,
  colback=gray!8,           
  colframe=black!20,        
  boxrule=0.5pt,
  width=0.5\textwidth,     
  title={\textbf{Few-shot Prompt Template}}
]
\footnotesize
\textbf{Header} \\
\textit{“Your task is to solve the following clinical multiple-choice question.”} \\[2pt]

\textbf{Question Block} \\
Present the target question text, then list answer options (A/B/C/\dots). \\[2pt]

\textbf{Instruction Block}
\begin{itemize}[leftmargin=*, noitemsep]
  \item The prompt shows $N$ solved \emph{reference examples}.
  \item Each example contains:
    \begin{itemize}[leftmargin=1em, noitemsep]
        \item \textbf{Finding Reasoning Paths}: brainstorming approaches
        \item \textbf{Reasoning Process}: a structured, step-by-step solution
    \end{itemize}
  \item Focus \emph{only} on learning the reasoning patterns.
  \item For the \emph{new} question, generate your \textbf{own} reasoning and answer.
  \item Constraints:
    \begin{itemize}[leftmargin=1em, noitemsep]
      \item Always choose one of the provided options—never “unknown”
      \item Follow the exact output format shown below
    \end{itemize}
\end{itemize}

\vspace{3pt}
\textbf{Output Format Hint} \\
\texttt{Thought: [your detailed step-by-step reasoning]} \\
\texttt{Answer: [one of A, B, C, \dots]} \\[6pt]

\textbf{Reference Examples} \\
For each retrieved example~$i$ insert: \\
\texttt{Reference Example $i$:} \\
\texttt{$<$question text$>$} \\
\texttt{$<$options$>$} \\
\texttt{Thought: $<$reference rationale$>$}
\end{tcolorbox}
\caption{Prompt template used for few-shot reasoning with retrieved reference examples.}
\label{tab:fewshot_prompt}
\end{table}

\section{The Reasoning Consistency Evaluator Rubric and Prompt}\label{A_Reasoning Consistency Evaluator Rubri}
To robustly aggregate multi-agent responses, we introduce a reliability scoring mechanism that evaluates the consistency between an agent’s reasoning and its final answer. In scenarios where a question has many answer options (e.g., $N$), simple majority voting becomes inefficient — achieving a reliable consensus typically requires at least $N+1$ agreeing agents.

To address this, we employ a scoring-based verification strategy: each agent’s reasoning is evaluated by a separate verifier agent that assigns a reliability score ranging from 1 to 5. This enables us to treat scores as soft confidence signals and aggregate responses more efficiently, even when only a few answers are available. The resulting per-sample reliability sum lies in the range of 4–20 (with 4 verifiers), providing fine-grained guidance for final answer selection. The full scoring prompt is shown in Table~\ref{tab:verifier_prompt}.

\begin{table}[t]
\centering
\begin{tcolorbox}[
  enhanced,
  sharp corners,
  colback=gray!8,
  colframe=black!20,
  boxrule=0.5pt,
  width=0.5\textwidth,
  title={\textbf{Reliability Scoring — System Prompt}}
]
\footnotesize
A clinical AI agent has answered the following multiple-choice question: \\[4pt]

\textbf{Question:} \\
\texttt{``\{question\}''} \\[6pt]

\textbf{Options:} \\
\texttt{``\{\$options\_str\$\}''} \\[6pt]

\textbf{The agent provided the following reasoning:} \\
\textbf{Thought:} \\
\texttt{``\{thought\}''} \\[6pt]

\textbf{Final Answer:} \\
\texttt{``\{answer\}''} \\[8pt]

\textbf{Your Role:} \\
You are a critical-thinking medical reviewer. Your task is to assign a reliability score from 1 to 5 based on how well the reasoning supports the answer. \\[6pt]

\textbf{Scoring Scale:}
\begin{itemize}[leftmargin=*, noitemsep]
  \item 5 — Reasoning is complete, medically accurate, and fully supports the answer.
  \item 4 — Mostly correct with minor issues, but the answer is still justified.
  \item 3 — Reasoning has some issues or omissions, but partially supports the answer.
  \item 2 — Reasoning is flawed or incomplete; answer is weakly supported.
  \item 1 — Reasoning is incorrect or misleading; answer is not justified.
\end{itemize}

\vspace{3pt}
\textbf{Output Format:} \\
\texttt{Score: [1-5]}
\end{tcolorbox}
\caption{System prompt used by the reasoning consistency evaluator to assign a soft reliability score.}
\label{tab:verifier_prompt}
\end{table}

\section{Reference CoT Dataset Examples}\label{app_refenrence_medreason}
We adopt the MedReason dataset~\cite{wu2025medreason} as our external reference corpus to support retrieval-augmented reasoning. MedReason comprises 32,682 high-quality question–answer pairs, each accompanied by detailed, clinically grounded chain-of-thought (CoT) explanations. The dataset is constructed through a knowledge graph–guided pipeline that ensures both logical consistency and medical factuality.

Specifically, the authors first collect QA pairs from seven public medical benchmarks, including MedQA, MedMCQA, PubMedQA, MMLU, MedXpert, Huatuo, and HLE. For each QA pair, relevant medical entities are extracted from both the question and the answer using GPT-4o and are then mapped to a structured medical knowledge graph, PrimeKG. Next, the shortest reasoning paths connecting the question and answer entities within the graph are retrieved and pruned using LLM-based selection to retain only clinically relevant paths. These paths serve as scaffolds for guiding step-by-step CoT generation.

To guarantee data quality, each generated reasoning trace is verified by prompting the LLM to reproduce the original answer solely based on the CoT explanation. If the answer cannot be recovered, the CoT is discarded. This quality filtering process reduces 45K generated samples to a final dataset of 32,682 verified CoTs, ensuring that each retained example is both logically sound and clinically accurate.

The 7,168th indexed sample from MedReason is shown below as an example: \\
\textbf{Question:}\\ 
A young male patient presents with complete rectal prolapse. The surgery of choice is? \\
\textbf{Answer Choices:}\\
A. Abdominal rectopexy \\
B. Delerom's procedure \\
C. Anterior resection \\
D. Goodsall's procedure \\
\textbf{Answer:}\\
Abdominal rectopexy. Explanation: Surgery is required, and the operation can be performed the perineal or the abdominal approaches. An abdominal rectopexy has a lower rate of recurrence, as an abdominal procedure risks damage to the pelvic autonomic nerves, resulting in possible sexual dysfunction, a perineal approach is also usually preferred in young men. Ref: Bailey \& Love’s Short Practice of Surgery, 25 ed., p. 1225\\
\textbf{Reasoning:}\\
Finding reasoning paths:
1. Rectal prolapse (disease) -> Surgical treatment options -> Perineal approach vs. Abdominal approach
2. Abdominal approach -> Abdominal rectopexy -> Risk of recurrence vs. risk of complications
3. Abdominal rectopexy -> Potential complications -> Pelvic autonomic nerve damage -> Male sexual dysfunction

Reasoning Process:
1.\textbf{Understanding Rectal Prolapse and Treatment Options}: Rectal prolapse is a condition where the rectum protrudes through the anus. Surgical intervention is often required, and the two main approaches are the perineal and abdominal approaches.

2.\textbf{Evaluating Surgical Approaches}:
-\textbf{Perineal Approach}: This is generally less invasive and may be preferred in older or frail patients due to lower surgical risk. However, it is associated with a higher rate of recurrence.
-\textbf{Abdominal Approach}: This includes procedures like abdominal rectopexy, which tend to have a lower recurrence rate. However, it is more invasive and carries a risk of complications.

3.\textbf{Abdominal Rectopexy and Complications}:
-\textbf{Recurrence Rate}: Abdominal rectopexy is favored for its lower recurrence rate compared to perineal procedures.
-\textbf{Complications}: A significant risk associated with the abdominal approach is potential damage to the pelvic autonomic nerves. This can lead to complications such as male sexual dysfunction, which is an important consideration in a young male patient.

Conclusion:
Abdominal rectopexy is the preferred surgical option for a young male patient with complete rectal prolapse due to its lower rate of recurrence. However, it is important to consider the risk of pelvic autonomic nerve damage, which could lead to male sexual dysfunction. This risk should be weighed against the benefits of a lower recurrence rate when deciding on the surgical approach. \\

\section{Test Dataset Examples}\label{app_test_data_eg}
\textbf{Test Datasets.} 
We conduct experiments on a curated benchmark comprising challenging subsets of nine medical QA datasets, selected from the MEDAGENTSBENCH framework~\cite{tang2025medagentsbench}. This benchmark is specifically designed to evaluate complex medical reasoning and includes questions that require multi-step clinical inference, diagnosis formulation, and treatment planning. The original datasets include:
\textbf{MedQA}~\cite{jin2021medqa}, a multilingual dataset with English questions derived from professional medical board exams (e.g., USMLE), focusing on diagnosis and treatment knowledge. Each question includes four answer choices, with an average length of 167 tokens.
\textbf{PubMedQA}~\cite{jin2019pubmedqa}, constructed from biomedical research articles in PubMed, requires reasoning over scientific evidence to determine yes/no/maybe answers. It contains long passages with an average length of 316 tokens.
\textbf{MedMCQA}~\cite{pal2022medmcqa}, sourced from Indian medical entrance exams (AIIMS and NEET-PG), comprises short multiple-choice questions across 21 medical subjects, averaging 18.7 tokens per question.
\textbf{MedBullets} \cite{chen2024medbullet}, a high-yield medical education platform, provides detailed clinical questions with up to five answer options and longer contexts (avg. 213 tokens).
\textbf{MedExQA} \cite{kim2024medexqa}, covers five medical specialties and emphasizes both correctness and explainability. It includes 935 questions, each with four options.
\textbf{MedXpertQA} \cite{zuo2025medxpertqa}, consists of two subsets: \textit{Reasoning} and \textit{Understanding}, specifically annotated to distinguish complex clinical inference tasks from knowledge recall.
\textbf{MMLU} \cite{hendrycks2020mmlu} and \textbf{MMLU-Pro} \cite{wang2024mmlu}, general-purpose multitask benchmarks that include medical subsets testing both factual recall and abstract reasoning.

To better reflect real-world difficulty, we follow the hard subset construction pipeline proposed by MEDAGENTSBENCH. Questions are selected based on model failure rates (<50\% accuracy across a set of strong models), medical topic coverage, and reasoning depth. Specifically, we include 100 hard questions each from MedQA, PubMedQA, MedMCQA, MedExQA, and MMLU-Pro; 100 from each MedXpertQA subset (Reasoning and Understanding); 89 from MedBullets; and 73 from MMLU. This results in a total of 862 expert-verified instances designed to stress-test the reasoning capabilities of large language models. Representative examples from the nine datasets are provided in Table~\ref{tab_test_nine_data}.

\begin{table*}[ht!]
\centering
\footnotesize
\renewcommand{\arraystretch}{1.3}  
\setlength{\tabcolsep}{6pt}        

\begin{tabularx}{\textwidth}{|p{1.7cm}|X|}

\hline
\rowcolor{gray!20}
\textbf{Dataset} & \textbf{Question} \& \textbf{Options} \& \textbf{Answer} \\
\hline

MedQA   & 
\textbf{Question}
A 39-year-old woman presents with headaches, constipation, increased thirst, and increased urinary frequency. Her blood pressure is 160/100 mm Hg and pulse is 67/min. Blood tests reveal high blood glucose and low potassium level. Abdominal CT scan shows a well-circumscribed homogeneous mass in the left adrenal gland. The hormone excess responsible for her symptoms uses which of the following pathways to exert its action? \textbf{Options}
(A): "Intracellular receptors", (B): "cAMP pathway", (C): "cGMP pathway", (D): "JAK/STAT pathway".
\textbf{Answer} (A) \\
\hline

PubMedQA  & 
\textbf{Question}
Sternal fractures in childhood are rare. The aim of the study was to investigate the accident mechanism, the detection of radiological and sonographical criteria and consideration of associated injuries.In the period from January 2010 to December 2012 all inpatients and outpatients with sternal fractures were recorded according to the documentation.A total of 4 children aged 5-14\u00a0years with a sternal fracture were treated in 2\u00a0years, 2\u00a0children were hospitalized for pain management and 2 remained in outpatient care.Sternal fracture in growing children : A rare and often overlooked fracture?\textbf{Options}
(A): "yes", (B): "no", (C): "maybe".
\textbf{Answer} (C) \\
\hline

MedMCQA   & 
\textbf{Question}
Minimum number of lobes require to form a tooth? \textbf{Options}
(A): "1", (B): "2", (C): "3", (D): "4".
\textbf{Answer} (C) \\
\hline

MedBullets  & 
\textbf{Question}
A 22-year-old woman presents to the emergency department with shortness of breath. She was hiking when she suddenly felt unable to breathe and had to take slow deep breaths to improve her symptoms. The patient is a Swedish foreign exchange student and does not speak any English. Her medical history and current medications are unknown. Her temperature is 99.5\u00b0F (37.5\u00b0C), blood pressure is 127/68 mmHg, pulse is 120/min, respirations are 22/min, and oxygen saturation is 90\% on room air. Physical exam is notable for poor air movement bilaterally and tachycardia. The patient is started on treatment. Which of the following parameters including forced expiratory volume in 1 second (FEV1), forced vital capacity (FVC), and diffusing capacity of carbon monoxide (DLCO) most appropriately describes this patient's underlying pathology? 
\textbf{Options}
(A): "Decreased airway tone", (B): "Increased FEV1", (C): "Increased FEV1/FVC", (D): "Increased FVC", (E): "Normal DLCO".
\textbf{Answer} (E)  \\
\hline

MMLU  & 
\textbf{Question}
How many different types of microorganisms may colonize the mouth? \textbf{Options}
(A): "35", (B): "100", (C): "350", (D): "500".
\textbf{Answer} (C) \\
\hline

MMLU-Pro  & 
\textbf{Question}
How are new polyomaviruses detailed? \textbf{Options}
(A): "Shot gun sequencing", (B): "Cultivation in human neural cells", (C): "Deep pyro sequencing (NGS)", (D): "Monoclonal antibody techniques".
\textbf{Answer} (A) \\
\hline

MedExQA  & 
\textbf{Question}
Which biological tissue has the highest viscosity?\textbf{Options}
(A): "Blood", (B): "Bone", (C): "Soft tissue", (D): "Water".
\textbf{Answer} (B) \\
\hline

MedXpert-R  & 
\textbf{Question}
A 52-year-old paralegal comes to the clinic reporting chronic low back pain that has gradually worsened over two years. She describes an achy pain rated at 7/10 and spends considerable time at her computer desk. She experiences numbness, tingling, and radiating pain down her right leg during prolonged sitting or standing. Her examination reveals:- No urinary/bowel incontinence or perineal numbness- Right lower extremity strength of 4/5- Decreased sensation in right leg- Right patellar reflex grade 1/4- Positive straight leg raise test on right sideWhich nerve root levels are most likely involved in this presentation? \textbf{Options}
(A) L1-L2 (B) L5-S1 (C) L2-L3 (D) S2-S3 (E) L4-L5 (F) L4-S1 (G) L3-L4 (H) S1-S2 (I) T12-L1 (J) L5-S2. \textbf{Answer}
G \\
\hline

 MedXpert-U  & 
Which hypoxic cell radiosensitizer, known for undergoing redox recycling or decomposing into a toxic product, has demonstrated effectiveness in the treatment of head and neck cancer? \textbf{Options} 
(A): Doranidazole, (B): Tirapazamine, (C): Camptothecin, (D): Misonidazole, (E): Pimonidazole, (F): Nimorazole, (G): Sanazole, (H): Cetuximab, (I): Etanidazole, (J): Methotrexate. \textbf{Answer}
(F) \\
\hline

\end{tabularx}
\caption{Dataset examples with corresponding questions and answer options from the nine test datasets.}
\label{tab_test_nine_data}
\end{table*}

\section{Fail case of existing method}
To illustrate the limitations of existing methods, we present a representative failure case from DeepSeek-R1 on a clinically complex MedQA question.
\textbf{Scenario:}  
A junior orthopaedic surgery resident is completing a carpal tunnel repair with the department chairman as the attending physician. During the procedure, the resident inadvertently cuts a flexor tendon, which is then repaired without complication. The attending advises the resident to exclude this incident from the operative report to avoid patient concern.\\[6pt]

\textbf{Question:}  
\textit{What is the correct next action for the resident to take?}\\[6pt]

\textbf{Answer Options:}
\begin{itemize}[leftmargin=1cm, label=--, itemsep=3pt]
    \item (A) Disclose the error to the patient and put it in the operative report
    \item (B) Tell the attending that he cannot fail to disclose this mistake
    \item (C) Report the physician to the ethics committee
    \item (D) Refuse to dictate the operative report
\end{itemize}

\textbf{Predicted Answer:} (A) - Disclose the error and document it.\\[8pt]

\textbf{Model's Reasoning:}
<think>
Okay, let's see. The question is about a resident who accidentally cuts a flexor tendon during carpal tunnel repair and the attending tells them not to report it. The options are A to D, and the generated solutions all say A.

First, the key points here are medical ethics, specifically honesty and patient disclosure. The resident made an error, even if it was repaired. The attending is advising to omit this from the operative report and not tell the patient to avoid worry.

In medical practice, complications should be documented and disclosed to the patient. The principle of informed consent requires patients to know about any complications, even if resolved. Failing to document could lead to legal issues and violates professional standards.

Option A says to disclose and document, which aligns with ethical guidelines. The other options either don't directly address informing the patient (like B, which is telling the attending but not necessarily documenting) or escalate without first handling it properly (C, reporting; D, refusing to dictate).

Since all solutions suggest A, the answer is likely A. They must inform the patient and include it in the report despite the attending's advice.
</think>
A

\noindent \textbf{Issue:} \\
The model selected \textbf{(A)} as the answer, which is inconsistent with the reference solution \textbf{(B)}.  
The error occurred because the model focused solely on disclosure and documentation, ignoring the specific instruction to address the attending's unethical guidance.

\end{document}